# A General Framework for Partial to Full Image Registration

Carlos Francisco Moreno-García & Francesc Serratosa
Robert Gordon University, School of Computing, Aberdeen, UK
Universitat Rovira i Virgili, Departament d'Enginyeria Informàtica i Matemàtiques,
Tarragona, Spain
c.moreno-garcia@rgu.ac.uk
francesc.serratosa@urv.cat

*Abstract*—Image registration is a research field in which images must be compared and aligned independently of the point of view or camera characteristics. In some applications (such as forensic biometrics, satellite photography or outdoor scene identification) classical image registration systems fail due to one of the images compared represents a tiny piece of the other image. For instance, in forensics palmprint recognition, it is usual to find only a small piece of the palmprint, but in the database, the whole palmprint has been enrolled. The main reason of the poor behaviour of classical image registration methods is the gap between the amounts of salient points of both images, which is related to the number of points to be considered as outliers. Usually, the difficulty of finding a good match increases when the image that represents the tiny part of the scene has been drastically rotated. Again, in the case of palmprint forensics, it is difficult to decide a priori the orientation of the found tiny palmprint image. We present a rotation invariant registration method that explicitly considers that the image to be matched is a small piece of a larger image. We have experimentally validated our method in two different scenarios; palmprint identification and outdoor image registration.

## 1 Introduction

Image registration is the process of transforming, comparing and integrating different sets of data into one coordinate system. Such data may be collected from multiple pictures, multiple points of view, or sensors working at different time lapses. It is used in computer vision, medical imaging, satellite data and image analysis in general. Interesting image registration surveys are [1] and [2], which explain the problematic of this goal. Certain areas of image registration in computer vision are interested in determining which parts of one image correspond to which parts of another image instead of searching for a whole correspondence. This problem often arises at early stages of applications such as scene reconstruction, object recognition and tracking, pose recovery and image retrieval. We define this specific type of image registration as "partial to full" image registration. The aim of this paper is to present a general method to solve this problem.

One of the most frequent uses of partial to full image registration is found on biometrics for forensic applications. In recent years, the use of palmprint recognition has increased with respect to fingerprint recognition in forensics applications [3], since the palm contains more features than the fingerprint, which makes the identification more feasible. Nevertheless, on crime scenes, it is more likely to find a small portion of the sample rather than the full palm. For these cases, partial to full image registration comes as a viable solution for this problem. Another current use of partial to full image registration is in applications that locate elements in outdoor scenes. By using only a small part of an image, for instance, a picture taken from a cell phone or an image obtained from social media, it is possible to find the location given a larger image, for instance, satellite or surveillance camera images. It is of basic importance to develop effective methods that are both robust in two aspects: being able to deal with noisy measurements and having a wide field of application.

Image registration methods are usually composed of two steps [4]. First, some salient points are selected from both images by using a feature extractor. Several methods have appeared to select salient points in images [5], for example SIFT [6], Harris corners [7] or SURF [8]. These methods are based on assigning some local features (for instance, a vector of 128 features) to each extracted point or pixel of the image. Each local feature usually depends on the information on the image given a radius and an angle. The second step is based on finding a correspondence between the extracted salient points or deducting the homography that transforms the coordinate system of one of the images to the other. Bipartite (BP) [9], or a new version called Fast Bipartite (FBP) [10] [11], is one of the most important algorithms used to find a correspondence between points or between graphs, especially if the second order relations between points are considered. This algorithm obtains the point correspondences but it does not deduct the homography, and it uses the features located at each point (for instance SIFT or SURF) or even the second order features (the relations between points). It is based on reducing the current problem to a linear assignation problem, and applying a linear solver such as the Hungarian method [12] or the Jonker-Volgenant solver [13]. Conversely, Iterative Closest Point (ICP) [14] is an algorithm employed to minimize the difference between two clouds of points. ICP is often used to reconstruct 2D or 3D surfaces from different scans. It only uses the position of the points but not the local features. Nevertheless, it has as advantage that it does not only obtain the correspondences, but a homography as well. It is usual to use ICP together with RANSAC [15], a method which discards points that do not fit on the deducted homography, thus eliminating the spurious correspondences. Those are considered as noise in the images or sensors. Most novel algorithms consider the features of each point and also the homographies, such as [16]. Finally, the Hough transform [17] [18] [19] is a technique used to find imperfect instances of objects represented by sub-sets of salient points within an image by a voting procedure. This voting procedure is carried out in a parameter space, from which object candidates are obtained as local maxima in an accumulator space that is explicitly constructed by the algorithm for computing the Hough transform.

In the literature, we find several examples that intend to solve the partial to full image registration problem, considering that the compared sets of salient points do not represent a unique coordinate system or set of characteristics. That is, they acknowledge that the salient points are grouped in several sets of points, since they assume different transformations or homographies are applied to each point set [20]. In [21], the method explicitly considered some different levels of occlusion and noise in the object's contour. Moreover, some methods have been presented which register several images at a time [22] to increase the probability of finding successful matches.

The main drawback of all of these methods is that their ability to obtain a trustworthy correspondence set strongly depends on the reliability of the tentative correspondences. In some image-registration based applications (forensic palmprint recognition, satellite images …), it is more usual to detect a tiny partial image rather than a full sample. In these cases, the tentative initial correspondences returned by the first step fail due to the great amount of outliers that have to be detected while comparing a tiny image to a full image. Thus, the second step (usually highly dependent on these initial correspondences) is not able to recover neither the correct correspondences nor the transformation matrix from the tiny image to the large one.

Previous work has been developed exclusively for partial to full palmprint identification in high-resolution images. These methods differ from the general partial to full image scenario mainly because of the methods to pre-process the image and the features that can be extracted from palms. While the general applications extract image features (e.g. corners [7], SIFT [6] or SURF [8]), in the case of palmprint, minutiae are extracted. The main features of minutiae are the type of minutiae (terminal or bifurcation) and the directional angle.

A first approach of partial to full palmprint matching was presented by Jain & Demirkus in [23]. The method consists of three major components: 1) Latent region of interest detection, which is only applied to the full palmprint image. 2) Feature extraction applied to both images. 3) Feature matching applied to both sets of features. Since it is generally know that the partial palmprints come from specific regions of the palm (i.e. thenar, hypothenar and interdigital), these regions are automatically detected, and features from these regions are utilized in the matching phase. The feature extraction phase obtains the SIFT features and the minutiae as well. In the feature matching phase, minutiae and SIFT matchers are used in parallel to obtain two different match scores; the score-based fusion is utilized to obtain the final match score. They reported an accuracy of 96% when performing a weighted fusion of minutiae and SIFT matching for synthetic partial palmprints of 500x500 pixels. When only using the minutiae features, the recognition rate was around 82%. Acknowledging that SIFT features could not be extracted from latent partial palmprints, Jain & Feng [24] presented a method exclusively for latent partial to full palmprint matching based in minutiae points and orientation field. The alignment is rigid and based on most similar minutiae pairs. Since latent palmprints are more difficult to match than synthetic palmprints, the accuracy decreased to 78.7% using larger partial palmprints (512x877 pixels in average).

In [25], Dai & Zhou presented a method based in minutiae points, ridge density, map, principal map and orientation field. The alignment is rigid and the matching is made through the Hough transform. Even though they achieve a recognition rate of 91,7% using synthetic partial palmprints of variable size, it is slow in computation. One year later, Dai *et. al* [26] presented a method which uses the average orientation field for coarse full palmprint alignment and the Generalized Hough Transform for fine segment level alignment, although it needs a manual alignment for partial palmprints. It has an accuracy of 91,9% with a slight improvement of the computational speed.

This paper presents a general method to perform partial to full image registration that we have called *PF-Registration*. We have applied different well-known techniques such as the Hough transform [17] [18], FBP [10] [11] and ICP [14]. Since it is not our interest to develop a method that only applies for palmprint applications we present, besides of our experiments on palmprints, a case of tiny area detection in large outdoor scenes.

The outline of our work is the following. In section 2, we describe our method and its computational complexity. In section 3, we validate our method using palmprint identification and outdoor scene recognition applications. Finally, we conclude the paper in section 4.

## 2 Partial to Full Image Registration

Consider we want to align an image $P$ that shows a tiny part of another image $F$. Both images are represented by their salient points, $(x^P, y^P) = \{(x_1^P, y_1^P), \ldots, (x_{|P|}^P, y_{|P|}^P)\}$ and $(x^F, y^F) = \{(x_1^F, y_1^F), \ldots, (x_{|F|}^F, y_{|F|}^F)\}$ together with some features extracted at the salient points $f^P = \{f_1^P, \ldots, f_{|P|}^P\}$ and $f^F = \{f_1^F, \ldots, f_{|F|}^F\}$. The number of salient points is $|P|$ and $|F|$, respectively.

The method we propose is based on two main steps. In the first step, $k$ positions $(x_1^c, y_1^c), \ldots, (x_k^c, y_k^c)$ on the full image $F$ are selected as candidates to be the centre of the partial image $P$. Moreover, the full image $F$ is split in sub-images $F_1, \ldots, F_k$, in which the centre of each image $F_a$ is the candidate position $(x_a^c, y_a^c)$. Each split image $F_a$ is represented by their set of salient points $(x^{F_a}, y^{F_a}) = \{(x_1^{F_a}, y_1^{F_a}), \ldots, (x_{|F_a|}^{F_a}, y_{|F_a|}^{F_a})\}$ and also their corresponding set of features $f^{F_a} = \{f_1^{F_a}, \ldots, f_{|F_a|}^{F_a}\}$. Due to the nature of the method, discrepancies may appear on the number $|P|$ of extracted salient points in the partial image and the number $|F_a|$ of extracted points in the split ones. Moreover, $|F| \leq \sum_{a=1}^{k}|F_a|$, since the split images can overlap.

In the second step, the algorithm seeks for the best alignment between the salient points $(x^P, y^P)$ of the partial image $P$ and the salient points $(x^{F_a}, y^{F_a})$ of each of the split images $F_1, \ldots, F_k$. To obtain these alignments, not only the salient point positions are used, but also their extracted features, more precisely, features $f^P$ and $f^{F_a}$. Thus, $k$ distances $D_1, \ldots, D_k$, together with $k$ correspondences $l_1, \ldots, l_k$ and $k$ alignments (also called homographies) $H_1, \ldots, H_k$ are computed. At the end of this second step, the method selects the image that obtains the minimum distance $D_{P,F}$ and returns the alignment $H_{P,F}$ and the correspondence $l_{P,F}$ between $P$ and $F$ that obtains this distance. On the following subsections we will explain in a deeper form each of the two steps of our *PF-Registration* method, which scheme is shown in Figure 1.

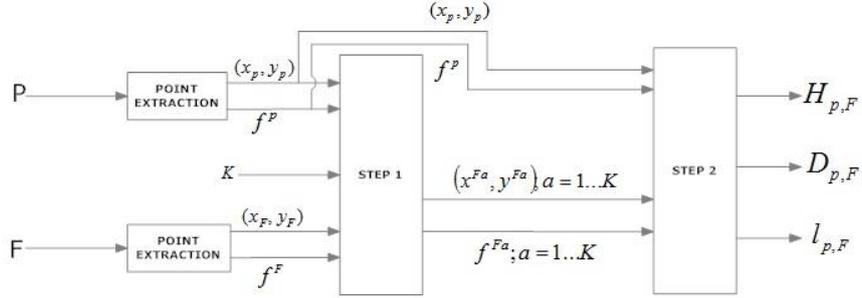

**Figure 1.** Diagram of the *PF-Registration* method.

## 2.1 Selecting position candidates

Figure 2 shows the main structure of the first step of our method. It is based on a Generalized Hough Transform [17], [18], [19]. As commented in the previous section, we assume the $|P|$ and $|F|$ salient points (positions and features) of both images have been previously extracted.

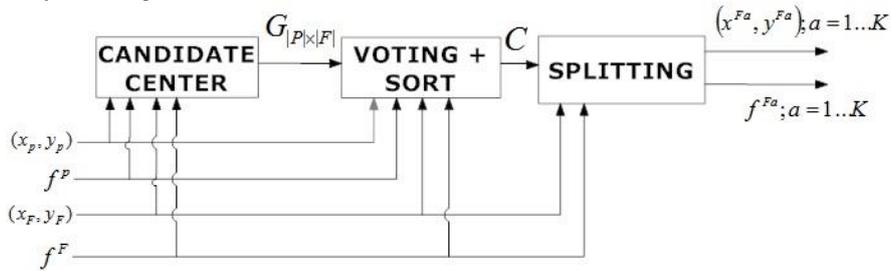

**Figure 2.** Diagram of Step 1.

The aim of the *Candidate Centre* module is to fill the $|P|x|F|$ matrix $G[i,j]$. Each cell of this matrix represents the position $(x_{ij}^C, y_{ij}^C)$ on the full image $F$ that the centre of the partial image $(\bar{x}, \bar{y})$ would obtain if the point $(x_i^P, y_i^P)$ on the partial image where mapped to the point $(x_j^F, y_j^F)$ on the full image. There are several forms to obtain these centres [19], which may use only one or several points, and also some information extracted from the features, such us angle information. The aim of this module is to detect the spatial relations on both images. If $s$ points in $P$ and $s$ points in $F$ have the same relative position, then $s$ cells of $G[i,j]$ will have the same value.

When matrix $G$ is filled, then the *Voting and Sorting* module generates an ordered list $C$ of the positions $(x_{ij}^C, y_{ij}^C)$ found in $G$, where $C = \{(x_1^c, y_1^c), \ldots, (x_T^c, y_T^c)\}$ through a clustering and voting process. List $C$ is set in a descendent order. That is, the positions with the most votes are the first ones. The voting process counts the number of centres, and also checks that their features are considered to be similar enough (distance smaller than a threshold $T_f$). To do so, an additional clustering process is performed that checks whether two centre points $(x_{ij}^C, y_{ij}^C)$ and $(x_{i'j'}^C, y_{i'j'}^C)$ have to be considered the same since they are close enough. That is, if their spatial distance is lower than a spatial threshold $T_s$. Thus, the *Voting and Sorting* module counts and orders the cells in G such that $dist^{position}\left((x_{ij}^C, y_{ij}^C), (x_{i'j'}^C, y_{i'j'}^C)\right) < T_s$ and $dist^{feature}(f_i^F, f_j^P) < T_f$. Both distances, that are application dependent, are normalised to be independent on the scale, rotation and global feature distortions.

Finally, with the best $k$ candidates to be the centre of the partial image on the full image, the set of points $(x^F, y^F)$ and the set of features $f^F$ are split in $k$ point sets $(x^{Fa}, y^{Fa})$, $1 \leq a \leq k$ and $k$ feature sets $f^{Fa}$, $1 \leq a \leq k$. Each point in $(x_i^F, y_i^F)$ is included in the set $F_a$ if $dist^{position}\left((x_i^F, y_i^F), (x_a^C, y_a^C)\right) \leq T_r$. The threshold $T_r$ represents the maximum radius of the set, meaning the maximum distance between any point and the centre of the set. Usually, it is determined depending on the radius of the partial set $(x^P, y^P)$. Parameter $k$ is application dependent and it is usually a trade-off between runtime and accuracy.

## 2.2 Best candidate selection through multiple correspondences

Figure 3 shows the second step of our registration method.

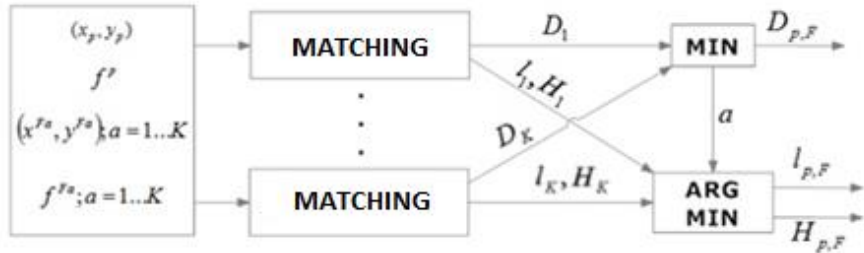

**Figure 3.** Diagram of Step 2.

In this second step, the method first seeks for the distances $D_a = dist(P, F_a)$; $1 \leq a \leq K$ and the correspondences $l_a$ between the points in each set, and also the homographies $H_a$ that transform $P$ to $F_a$. Several algorithms can be used to find these correspondences and homographies. These algorithms use the positional information $(x^{Fa}, y^{Fa})$ and $(x^P, y^P)$ and also their features $f^{Fa}$ and $f^P$. For example, the Hungarian method [12] or its upgrade [13], the ICP [14] (where no outliers are considered), the RANSAC method [15] that considers the presence of outliers, the Bipartite Graph Matching [9],[10], [11]that considers second order information, and more sophisticated ones [16]. Even a greedy algorithm that simply selects the best option without considering the other candidates could be used.

Given the data obtained from the *Matching* module, we wish to select the set of points $(x^{Fa}, y^{Fa})$ that obtained the minimum distance $D_a$. This is because we assume $D_a$ is agood enough approximation of$dist(P, F)$. Moreover, we also assume the correspondence an alignment (homography) between $P$ and $F$ approximates the correspondence $l_a$ and homography $H_a$. Therefore, $l_{P,F} = l_a$ and $H_{P,F} = H_a$.

To conclude, breaking down the full image into a set of candidates instead of performing straightforward partial to full comparison has two important advantages. On the one hand, the computational cost of obtaining the $k$ distances $D_a$ is lower than obtaining directly the value $D_{P,F}$. On the other hand, our method obtains a more precise local minimum, since we only use a certain amount of salient points for each match. In the next sub-section, we discuss the aforementioned statements.

### 2.3 Computational Complexity

In this section we show the computational complexity of each module of our method. The first step of the method is the *Candidate Centre* module, where we seek for $k$ candidate positions where the partial is found over the full image. The complexity cost of this module is $O(|P| \cdot |F|)$, since the calculation of matrix $G$ depends solely on the number of salient points extracted in partial image $P$ and full image $F$. Afterwards, the *Voting and Sorting* module generates a list of the centres according to the voting frequency. Once again, an $O(|P| \cdot |F|)$ complexity is derived, since this process requires to find the best $k$ candidates within matrix $G$. At the end of step 1, we encounter the *Splitting* module, where $k$ partial images $F_{1,\ldots,K}$ are created from the full one. The complexity of this module is $O(k \cdot |P|)$, since this process is executed $k$ times, and each $F_a$ has, on average, $|P|$ salient points.

On the second step, we defined the *Matching* and *Minimisation* modules. The first one involves the use of a matching method to find the correspondence, the homography and the distance between the tiny image and the $K$ partial images derived from $F$. The module's computational complexity is dependent of the selected matching method. For example, the complexity of using the Hungarian method is $O(k \cdot (|P|)^3)$. Meanwhile, the *Minimisation* module has a constant computational complexity since the minimum distance $D_a$ with its corresponding $l_a$ and $H_a$ is selected in parallel with the previous module.

All in all, the highest computational complexity of our method depends on the *Matching* module. If we use the Hungarian method (one of the simplest methods) the complexity is $O(k \cdot (|P|)^3)$. The computational complexity of a classical registration depends on the matching method, and so it is $O((|F|)^3)$ if the Hungarian method is used. Since it is usual to have a small value of $k$ (for instance equal or lower than 4) and considering we assume $|P| \ll |F|$, then $O(k \cdot (|P|)^3) \ll O((|F|)^3)$. This inequality makes us realise that our method has an important speedup with respect to a general image registration method.

## 3 Experimental validation

We have experimentally validated our method on two distinct databases, which are a palmprint database and an outdoors database. Moreover, we have tested different matching algorithms for the *Matching* module. In both experiments, we considered $k = 4$ (number of tentative centres).

### 3.1 Matching Method Selection through a Palmprint Database

To evaluate the efficiency of the *PF-Registration* method in a standalone situation, the only modules throughout our method that we have to define are the point extractor and the matching algorithm. The other modules are independent of the application. On the one hand, we used the extractor presented in [25], [26] to obtain the salient points from each image (both in the partial and the full images). On the other hand, we used the Hough method proposed in [28], the Iterative Closest Points method (ICP) [12] and the Fast Bipartite algorithm (FBP) [10] as the matching algorithms to be compared. The first method was considered since it is able to work with few salient points and it is a state of the art method used for matching. Nevertheless, it has not the ability to reject outliers, thus we are interested in seeing if the difference between its performance and the other two algorithms is significant. The second algorithm, ICP, was selected since it is a fast and general method used to match salient points. Finally, the FBP algorithm was selected since it is a fast and efficient matching algorithm, which is capable to deal with outliers.

We used images contained in the Tsinghua 500 PPI Palmprint Database [25], [26]. It is a public high-resolution palmprint database composed of 500 palmprint images of 2040 x 2040 resolution and captured with a commercial palmprint scanner from Hisign. We selected the first 10 subjects of the database [27]. From each of these subjects, 8 images of the same person are enrolled. Then, we considered the first four palmprints belong to the reference set, and the last four belong to the test set. Therefore, the full palmprints are the same as the reference set (figure 4.a) and the partial palmprints are a circular patch of a palmprint on the test set (figure 4.b). The partial palmprints have been extracted given a variable radius from 0.5 to 2.5 cms and a random minutia as the centre of the patch. Full palmprints have an average of 800 minutiae.

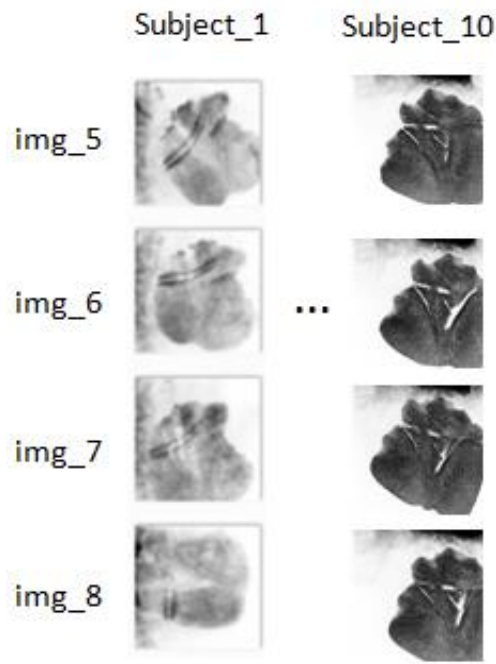

**Figure 4.a** Some images from the reference set, which is composed of 40 palmprints.

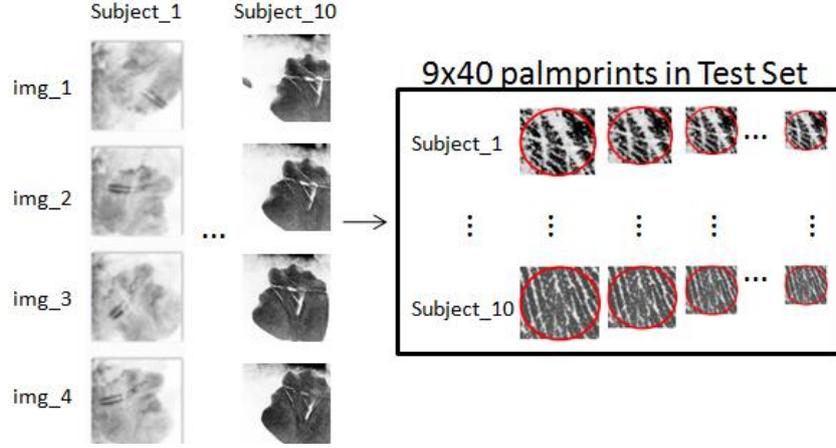

**Figure 4.b** Test set composed of 9x40 patches of different radius extracted from full palmprints.

To summarise, we have 40 full images on the reference set and other 40 in the test set. From the latter set, by cutting the 9 circular patches (radius 0.5,0.75,…,2.25,2.5 centimetres), we create a total of 360 partial palmprints. We compare every circular patch with every image on the reference set, thus computing a total of 40x40x9=14'440 comparisons (1'600 per radius). This process is repeated per each of the three matching algorithms.

Following the same nomenclature as section 2, two minutiae $m_i^P$ and $m_j^{F_a}$ extracted from palmprints $P$ and $F_a$ are represented by positions $(x_i^P, y_i^P)$ and $(x_j^{F_a}, y_j^{F_a})$ and two features $f_i^P = (\theta_i^P, t_i^P)$ and $f_j^{F_a} = (\theta_j^{F_a}, t_j^{F_a})$. Feature $\theta_i$ represents the directional angle of the ridge at the minutia point, and $t_i$ represents the type of minutia (termination or bifurcation) [17]. If both minutiae belong to the same type then $t_i^P = t_j^{F_a}$ and the distance is defined as $dist^{feature}(f_i^{F_a}, f_j^P) = cyclical\_dist(\theta'^{F_a}_i, \theta'^P_j)$. Otherwise, these minutiae cannot be mapped and so, $dist^{feature}(f_i^{F_a}, f_j^P) = \infty$. Minutiae angles have been normalised with respect to the average angle to be, to the most, independent of the rotations, $\theta'^P_i = \theta_i^P - \bar{\theta}^P$ and $\theta'^{F_a}_i = \theta_i^{F_a} - \bar{\theta}^{F_a}$.

The distance between positions is defined as $dist^{position}\left((x_i^P, y_i^P),(x_j^{F_a}, y_j^{F_a})\right) = Euclidean\_dist\left((x'^P_i, y'^P_i),(x'^{F_a}_j, y'^{F_a}_j)\right)$, where $(x'^P_i, y'^P_i)$ is the position of minutiae $(x_i^P, y_i^P)$ in which a translation to the centre of the partial $(x^C, y^C)$ and also a rotation has been applied. The angle of this rotation is the average angle $\bar{\theta}^P$ of the minutiae in the partial palmprint. Similarly, $(x'^{F_a}_j, y'^{F_a}_j)$ is the position of $(x_j^{F_a}, y_j^{F_a})$ translated to the centre of $F_a$. A rotation of the mean angle $\bar{\theta}^{F_a}$ has been applied as well. That is, $(x'^P_i, y'^P_i) = rotate_{\bar{\theta}^P}\left((x_i^P, y_i^P) - (x^C, y^C)\right)$ for the partial and $(x'^{F_a}_j, y'^{F_a}_j) = rotate_{\bar{\theta}^{F_a}}\left((x_j^{F_a}, y_j^{F_a}) - (x_a^C, y_a^C)\right)$ for the full.

The distance between two minutiae $m_i^P$ and $m_j^{F_a}$ is defined as follows

$$dist(m_i^P, m_j^{F_a}) = w^{feature} \cdot dist^{feature}(f_i^{F_a}, f_j^P) + w^{position} \cdot$$
$$dist^{position}\left((x_i^P, y_i^P), (x_j^{F_a}, y_j^{F_a})\right)$$
(1)

where weights $w^{feature}$ and $w^{position}$ depend on the data. Finally, given two partial palmprints, the distance between them is defined as follows:

$$D(P, F_a) = \min_{\forall l_a} \frac{\sum_{m_i^P} dist(m_i^P, m_{l_a(i)}^{F_a})}{|P|}$$
(2)

Figure 5 shows the recognition ratio of our method given the three selected matching algorithms respect to the patch radius (in centimetres). We have used the 3-nearest neighbours classification algorithm. Independently of the radius, FBP obtains the best results. Note the recognition ratio of FBP and ICP tends to increase when the radius increases but this is not the case for the Hough algorithm, which seems to keep the same recognition ratio for radius larger than 1.5 cms.

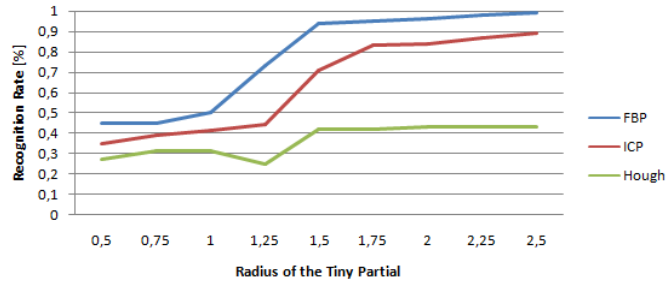

**Figure 5.** Percentage of properly classified patches with respect to the radius (cms)

Figure 6 shows the average runtime (in hours) to compare one tiny patch to the full reference set. We performed these tests using a PC with Intel 3.4 GHz CPU and Windows 7 operating system. As the radius increases, the gap between the runtime of these algorithms increases as well.

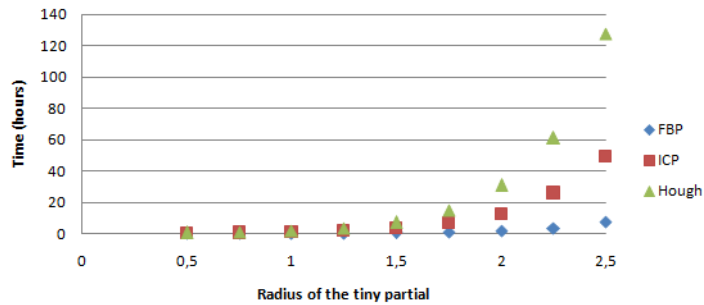

**Figure 6.** Runtime to compare a patch in the whole database (in hours) with respect to the radius in (cms).

A specific tiny to full palmprint matching was reported in [17]. They used the same database as we did and the experimental validation shows a recognition ratio of

82% given an approximate radius of 1.75 cms (in fact, they used square patches instead of circular ones). They also report that their method does not obtain acceptable results given smaller patches. Note our method obtains a recognition ratio of 92% using FBP and 83% using ICP given the same radius. Therefore, although our method is not palmprint oriented, we obtain similar or higher recognition ratios.

### 3.2 Image Recognition in Outdoor Scenes

In this second experimentation, we used images contained in the "Sagrada Familia" database 0. It consists of 364 pictures of 480x718 pixels taken from all around the Sagrada Familia cathedral in Barcelona, Spain. These images are stored in a sequential order, thus knowing that the first image was taken right next to the second image in terms of proximity, and so on. We extract the 800 most important features from the top half of each image using SURF [8].The reference set is composed of the odd numbered images (figure 7.a). The test set is composed of patches extracted from the even numbered images. More precisely, from each even image, we selected 9 circular patches with radius from 20 to 100 pixels. The centres of the patches are the centres of masses of the salient points of the original images (figure 7.b).

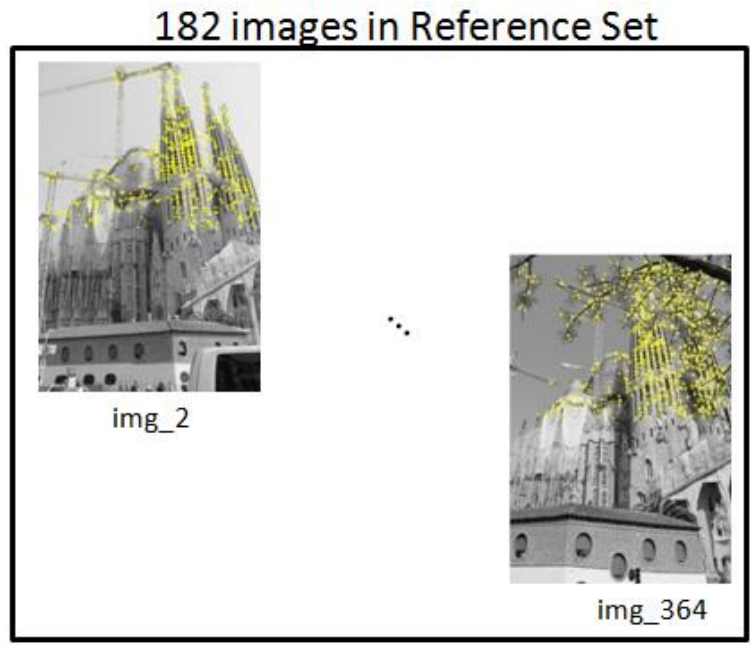

**Figure 7.a** Two examples of the reference set images with the extracted salient points.

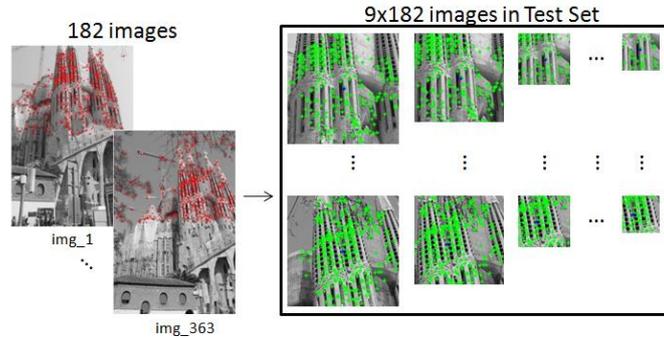

**Figure 7.b** Two examples of the test set images with the extracted salient points and the generated patches.

We compared every circular patch in the test set to every image on the reference set, thus computing a total of 182x182x9=298'116 correspondences in total (33'124 per radius). The FBP algorithm was used in the *Matching* module since it obtained the best results in the first experiments.

Figure 8.a shows the obtained distances of all combinations of the reference set (182 original images represented as columns) and the test set (182 patches of radius 2.5 represented as rows). Although the diagonal of the matrix in general has lower distance values, clearly the distance cannot be considered as a good metric to classify these patches. This is because there is a large variability on the number of salient points in the patches. In a similar way, figure 8.b shows the number of mappings between patches and images. FBP algorithm has the ability to discard outliers. Thus, figure 8.b shows the number of inliers considered by the FBP algorithm. In this case, the diagonal seems to be more visible since it tends to have more inliers than the rest of the cells. Finally, figure 8.c shows the obtained distance (figure 8.a) normalised by the number of inliers (figure 8.b). That is, we want to know the quality of the obtained mappings and we do not want the number of outliers to influence on the metric value. In this final metric, the diagonal clearly tends to have lower values than the rest of the cells.

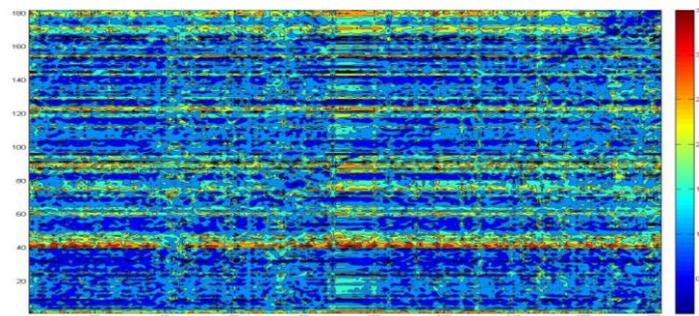

a)   Distance matrix given the reference set and test set with radius 2.5 cms.

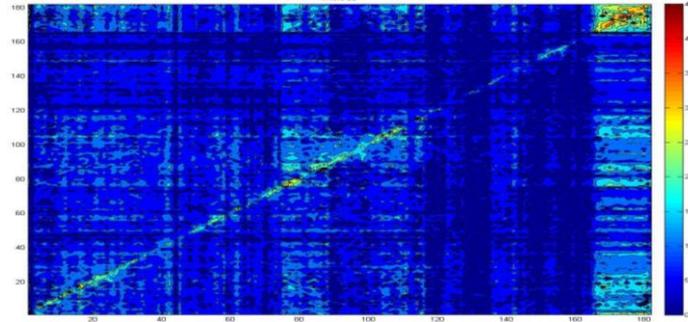

b) Number of inliers between the reference set and test set with radius 2.5 cms.

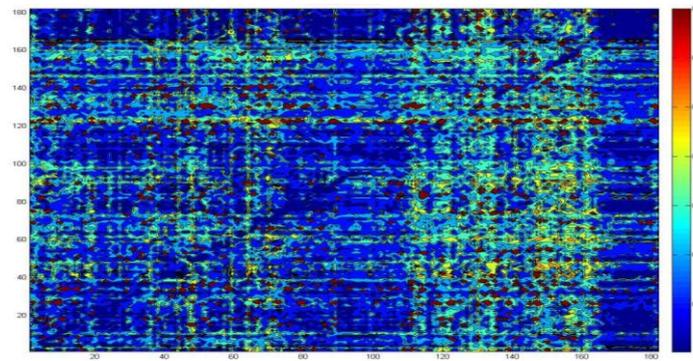

c) Distance normalised by the number of inliers given the reference set and test set with radius 2.5 cms.

**Figure 8.** Three different metrics used to evaluate the quality of the tiny to full matching algorithm.

Since images are ordered in the database considering the camera has been moving around the cathedral, images from the test set that are more similar to the reference set are the ones that are closer in its order. For this reason, in this experiment, we want to know how many times the system is able to identify a neighbour of the image (considering the order) as the one that obtains the minimum distance. We used the distance normalised by the number of inliers as a metric and the FBP as the matching algorithm. Figure 9 shows the ratio of images that the closest one is one of their neighbours and figure 10 shows the runtime in seconds to compute the comparison of one partial image against the whole reference set. We used a PC with Intel 3.4 GHz CPU and Windows 7 operating system. The insertion and deletion costs used on the FBP method were 0.1, and the whole features were normalized.

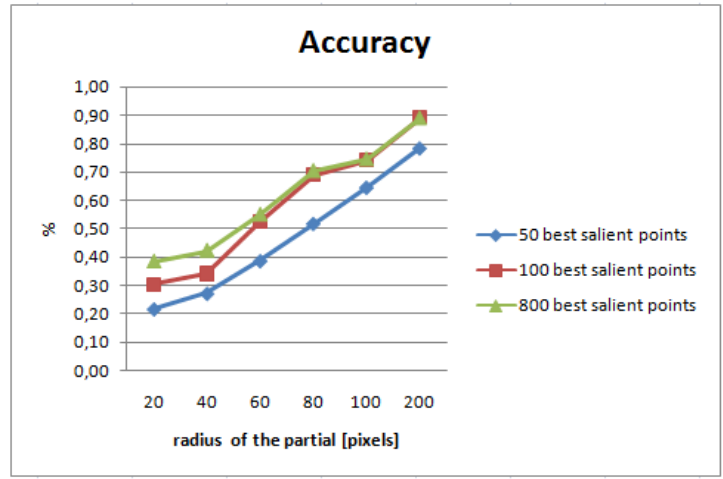

**Figure 9.** Ratio of images that the minimum distance is achieved by one of their neighbours.

| radius [px] | 20 | 40 | 60 | 80 | 100 | 200 |
|---|---|---|---|---|---|---|
| 50 best salient points | 853,41 | 876,85 | 902,64 | 904,24 | 910,34 | 910,39 |
| 100 best salient points | 917,64 | 921,18 | 925,65 | 946,02 | 994,32 | 1065,70 |
| 800 best salient points | 951,37 | 988,78 | 1076,91 | 1999,37 | 2702,01 | 15458,47 |

**Figure 10.** Runtime in second to compare a patch against the whole reference set.

Notice that there is practically no difference in accuracy between using the best 100 salient points instead of the whole 800 salient points in the available dataset. Nevertheless, in terms of runtime, the difference is drastically larger. Considering that the random chance of obtaining the correct result is $\frac{2}{182} \cong 0,11$ and that the sample sizes are tiny (e.g. a 20 pixel radius derives in approximately 5-10% of the full image), we believe that our reported accuracy is good when using a high quality and real image database with no previously established oracle. Besides, our tiny to full matching is performed between two images taken from different perspectives.

## 4 Conclusions

Several methods have been presented to solve image registration. Some of them are general methods applicable to a large spectrum of problems but other ones are application dependent. In some cases, image registration is based on finding a tiny patch of the image into a larger one. In these cases, most of the methods (general methods or application dependent ones) that do not consider this specific feature do not obtain optimal results. This paper presents a non-application dependent method that specifically considers the case that one of the images is a tiny part of the other one. It is based on two main steps that involve a matching and voting process.

In the experimental section, we have shown the functionality of the method in two completely different applications, palmprint recognition and outdoor scenes detection.

We have deducted the Fast Bipartite Graph is the matching algorithm that has achieved a higher accuracy in a reduced runtime. In the specific palmprint experiments, we have compared our method to another one that it was designed to specifically compare tiny to full palmprints. Results show that we obtain similar accuracies although our method can be applied to a larger set of frameworks.

## Acknowledgements

This research is supported b the CICYT project DPI2013-42458-P, by project TIN2013-47245-C2-2-R and Consejo Nacional de Ciencia y Tecnología (CONACyT Mexico).

## References

[1] Salvi, J., Matabosch, C., Fofi, D., Forest, J., "A review of recent range image registration methods with accuracy evaluation". Image Vision Comput. 25(5): 578-596, 2007.

[2] Ardeshir, Goshtasby, A., "2-D and 3-D image registration for medical, remote sensing, and industrial applications". Wiley Press, 2005.

[3] Dewan, S.K., "Scan a Palm, Find a Clue". The New York Times. W. Elementary. November 2003.

[4] Zitová, F., "Image registration methods: a survey". Image Vision Com. 21(11): 977-1000, 2003.

[5] Mikolajczyk, K., Schmid, C., "A performance evaluation of local descriptors", IEEE Trans. Pattern Anal. Mach. Intell. 27 (10), pp: 1615–1630, 2005.

[6] Lowe, D.G., "Distinctive image features from scale-invariant keypoints". IJCV 60(2), pp. 91–110. 2004.

[7] Harris, C., Stephens, M., "Proceedings of the 4th Alvey Vision Conference". pp. 147–151, 1988.

[8] Bay, H., Ess, A., Tuytelaars, T., Van Gool, L., "SURF: speeded up robust features". Computer Vision and Image Understanding (CVIU), Vol. 110, No. 3, pp. 346—359, 2008.

[9] Riesen, K., Bunke, H., "Approximate graph edit distance computation by means of bipartite graph matching". Image Vision Comput. 27(7): 950-959, 2009.

[10] Serratosa, F., "Fast computation of bipartite graph matching". Pattern Recognition Letters, PRL 45, pp: 244–250, 2014.

[11] Serratosa, F., "Speeding up fast bipartite graph matching trough a new cost matrix". International Journal of Pattern Recognition and Artificial Intelligence 29 (2), 2015.

[12] Kuhn, H.W., "The hungarian method for the assignment problem export". Naval Research Logistics Quarterly 2(1-2), 83–97, 1955.

[13] Jonker, R., Volgenant, T., "Improving the hungarian assignment algorithm". Operations Research Letters. Vol. 5 Issue 4, pp. 171-175, 1986.

[14] Zhang, Z., "Iterative point matching for registration of free-form curves and surfaces", Int. J. Comput. Vision 13 (2), pp: 119–152, 1994.

[15] Fischler, M.A., Bolles, R.C., "Random sample consensus: a paradigm for model fitting with applications to image analysis and automated cartography". Commun. ACM 24 (6), pp: 381–395, 1981.

[16] Sanromà, G., Alquézar, R., Serratosa, F., Herrera, B., "Smooth point-set registration using neighbouring constraints". Pattern Recognition Letters 33, pp: 2029-2037, 2012.

[17] Jain, A.K., Flynn, P., Ross, A.A., "Handbook of Biometrics". Springer. 2009.

[18] Ballard, D.H., "Generalizing the hough transform to detect arbitrary shapes". IEEE Trans. on Pattern Analysis and Matching Intelligence, 1980.

[19] Kassim, A.A., Tan, T., Tan, K.H., "A comparative study of efficient generalised hough transform techniques". Image and Vision Computing 17, pp: 737–748, 1999.

[20] Kokiopoulou, E., Frossard, P., "Graph-based classification of multiple observation sets". Pattern Recognition 43, pp: 3988–3997, 2010.

[21] Chang, L., Arias-Estrada, M., Hernández-Palancar, J. Enrique Sucar, L., "Partial shape matching and retrieval under occlusion and noise". CIARP 2014, LNCS 8827, pp. 151-158. Springer. 2014.

[22] Wachinger, C., Navab, N., "Simultaneous registration of multiple images: similarity metrics and efficient optimization", IEEE Trans. on Pattern Analysis and Matching Intelligence 35 (5), pp: 1221-1233, 2013.

[23] Jain, A.K., Demirkus, M., "On latent palmprint matching". MSU Technical Report, 2008.

[24] Jain, A.K., Feng, J., "Latent palmprint matching". IEEE Trans. on PAMI, 2009.

[25] Dai, J. Zhou, J., "Multifeature-based high-resolution palmprint recognition". IEEE Trans. Pattern Analysis and Machine Intelligence 33 (5), pp: 945-957, 2011.

[26] Dai, J., Feng, J., Zhou, J., "Robust and efficient ridge based palmprint matching". IEEE Transactions on Pattern Analysis and Matching Intelligence 34 (8), 2012.

[27] http://deim.urv.cat/~francesc.serratosa/databases/

[28] Ratha N.K., Karu K., Chen S., Jain A.K., "A real-time matching system for large fingerprint databases," IEEE Trans. on PAMI 18 (8), pp. 799−813, 1996.

[29] Penate-Sanchez, A., Moreno-Nogue, F., Andrade-Cetto, J., Fleuret, F., "LETHA: learning from high quality inputs for 3D pose estimation in low quality images". Proceedings of the International Conference on 3D vision, 2014.